\documentclass[letterpaper, 10 pt, journal, twoside]{IEEEtran}
\usepackage{amsmath}
\usepackage{amssymb}
\usepackage[font = small]{caption}
\usepackage{subcaption}
\usepackage{dblfloatfix} 
\usepackage{tikz}
\usepackage{tikz-3dplot}
\usepackage[hidelinks]{hyperref}
\usepackage{orcidlink}
\usepackage[english]{babel}
\newcommand\B{\textbf}
\newcommand\I{\textit}
\begin{document}
\title{A PnP Algorithm for Two-Dimensional Pose Estimation}
\author{Joshua Wang \orcidlink{0009-0006-3221-5825}
}
\maketitle
\begin{abstract}
We propose a PnP algorithm for a camera constrained to two-dimensional motion (applicable, for instance, to many wheeled robotics platforms). Leveraging this assumption allows accuracy and performance improvements over 3D PnP algorithms due to the reduction in search space dimensionality. It also reduces the incidence of ambiguous pose estimates (as, in most cases, the spurious solutions fall outside the plane of movement). Our algorithm finds an approximate solution by solving a polynomial system and refines its prediction iteratively to minimize the reprojection error. The algorithm compares favorably to existing 3D PnP algorithms in terms of accuracy, performance, and robustness to noise.
\end{abstract}

\section{Introduction}
\IEEEPARstart{T}{he} perspective-$n$-point (PnP) problem asks us to find the pose of a perspective camera using correspondences between image points and points in world coordinates. We are given $n$ 3D points $\B{p}_{i}$ and their corresponding projections onto the image plane, $\B{q}_{i}$. We wish to find a rigid transformation matrix $\B{R}$ from the camera frame to the world frame such that, for each $1\leq i\leq n$:

$$z_{i}\begin{bmatrix}\B{q}_{i}\\ 1\end{bmatrix} = \B{I}_{3\times 4}\B{R}^{-1}\begin{bmatrix}\B{p}_{i}\\ 1\end{bmatrix}$$

Here, the $z_{i}$ are the depth factors and we use known camera intrinsics to remap the $\B{q}_{i}$ so that they model the normalized perspective camera. This is exactly solvable in the case $n=3$, giving rise to the class of P3P algorithms as in \cite{P3P}. P4P and P5P algorithms have also been developed in this vein \cite{P4P}. However, using all available correspondences is desirable to reduce the impact of noise, in which case the problem is overconstrained and we must settle for an approximate solution. One approach is to minimize the \I{reprojection error}, the sum of squared distances between each $\B{q}_{i}$ and the projection of its $\B{p}_{i}$ onto the predicted image plane. This is given by:

$$\text{Error}(\B{R}) = \sum_{i}\left\Vert\begin{bmatrix}\B{q}_{i} \\ 1\end{bmatrix}-\frac{1}{z_{i}}\B{I}_{3\times 4}\B{R}^{-1}\begin{bmatrix}\B{p}_{i} \\ 1\end{bmatrix}\right\Vert^{2}$$

Note that this does not necessarily measure the distance in \I{pixels} due to our normalization of the $\B{q}_{i}$. Minimization of the reprojection error is a simple, geometric criterion commonly used in iterative PnP solutions; there are many alternative errors, such as the object-space error of Lu et al. \cite{LHM}. Non-iterative solutions also exist, such as the EPnP algorithm presented by Lepetit et al., which utilizes four weighted control points to reduce time complexity to $O(n)$ \cite{EPnP}, or the OPnP algorithm of Zheng et al. which analytically minimizes an algebraic error by solving a system of polynomial equations \cite{OPnP}. At the time of publication, state-of-the-art PnP solutions were mostly non-iterative polynomial system solvers, but iterative solutions such as Lu et al.'s LHM still remain competitive, and many non-iterative solvers, including \cite{EPnP,OPnP}, have an iterative phase at the end to refine the predicted pose (for this purpose, a simple optimization algorithm like Gauss-Newton suffices).

PnP algorithms have excellent accuracy for large numbers of correspondences spread widely in 3D space, but many behave suboptimally when the points are coplanar. Some algorithms such as EPnP include a special case for planar configurations, and assuming coplanarity allows the use of homography-based solvers including the IPPE solver of Collins and Bartoli \cite{IPPE}, which take advantage of the problem's unique structure to achieve higher precision. However, planar configurations present a challege even for specialized solvers because they can produce ambiguity; the images seen from multiple camera positions can be very similar (or even indistinguishable given the presence of noise). The problem worsens at low resolutions and long distances, or when the plane containing the points is almost parallel to the image plane. We illustrate the possible ambiguity with a square fiducial marker, the corners of which are often used for pose estimation:

\begin{figure}[h]
    \centering
    \captionsetup{justification=centering}
    \includegraphics[width=0.75\linewidth]{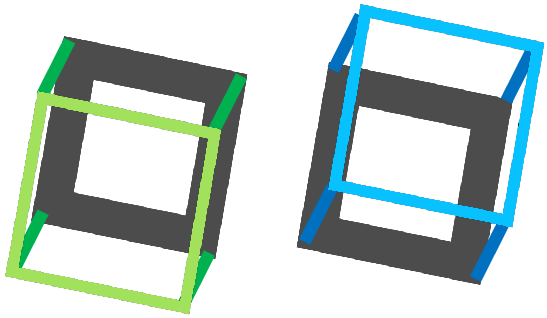}
    \caption{Ambiguity in a square's estimated pose \cite{Planar1}}
\end{figure}

This problem is well-known \cite{Planar1, Planar2}, and many methods have been proposed to mitigate it. For instance, iterative algorithms can be initialized with the last known pose to improve stability (although this means that they can become persistently stuck in the wrong pose). Some algorithms, such as IPPE, attempt to compute all possible solutions, either selecting the best fit or returning the solutions to the user. The outputs of existing algorithms can be fused with other sensor information as in \cite{EKF}. However, as robust as these solutions are, they solve only a facet of the deeper issue of imprecision for planar configurations.

This paper deals with the case of a camera constrained to two-dimensional movement (for instance, mounted to a wheeled robot). Here, PnP algorithms are commonly used to estimate pose from environmental landmarks or fiducials. However, estimating full 3D pose loses valuable information in this situation, as the returned 3D poses must be projected onto the plane of movement. An algorithm which takes the 2D constraint into account in the first place would likely exhibit improved accuracy because it can account for error while remaining in the plane of motion instead of `escaping' into disallowed poses. Performance improvements could also be realized because of the reduction of the search space's dimensionality. We derive an example of such an algorithm, called \B{2DPnP}, which succeeds in these criteria despite its simplicity. Our algorithm finds an approximate solution using combined geometric and algebraic methods, which it uses as a starting point to iteratively minimize the reprojection error.
\section{2DPnP}
We derive an expression for $\text{Error}(\B{R})$ to be used in the iterative phase of the algorithm. Assume that the camera origin is on the world $xy$-plane and that the camera is constrained to move in this plane. Then, $\B{R} = \B{P}\begin{bmatrix}\B{C} & \B{0} \\ \B{0}^{\intercal}& 1\end{bmatrix}$ where $\B{P}$ is an unknown transformation preserving the $xy$-plane and $\B{C}$ is a known rotation matrix representing the camera's angular offset. Let $\B{C} = \B{R}_{\alpha}\B{R}_{\beta}\B{R}_{\gamma}$ be $\B{C}$'s decomposition into the $ZYZ$ Euler angles $\alpha,\beta,\gamma$. Note that $\B{R}_{\alpha}$ can be ``absorbed" into $\B{P}$, so we may assume that $\alpha = 0$. Then, we have:

$$\frac{1}{z_{i}}\B{I}_{3\times 4}\B{R}^{-1}\begin{bmatrix}\B{p}_{i}\\ 1\end{bmatrix} = \frac{1}{z_{i}}\B{I}_{3\times 4}\begin{bmatrix}\B{R}_{\gamma}^{-1}\B{R}_{\beta}^{-1} & \B{0} \\ \B{0}^{\intercal} & 1\end{bmatrix}\B{P}^{-1}\begin{bmatrix}\B{p}_{i}\\ 1\end{bmatrix}$$
$$= \frac{1}{z_{i}}\B{R}_{\gamma}^{-1}\B{I}_{3\times 4}\begin{bmatrix}\B{R}_{\beta}^{-1} & \B{0} \\ \B{0}^{\intercal} & 1\end{bmatrix}\B{P}^{-1}\begin{bmatrix}\B{p}_{i}\\ 1\end{bmatrix}$$

Note that multiplying by the rotation $\B{R}_{\gamma}$ does not change $z_{i}$ because it rotates about the camera's axis. Thus:

$$\left\Vert\B{R}_{\gamma}\begin{bmatrix}\B{q}_{i} \\ 1\end{bmatrix} - \frac{1}{z_{i}}\B{I}_{3\times 4}\begin{bmatrix}\B{R}_{\beta}^{-1} & \B{0} \\ \B{0}^{\intercal} & 1\end{bmatrix}\B{P}^{-1}\begin{bmatrix}\B{p}_{i}\\ 1\end{bmatrix}\right\Vert$$
$$=\left\Vert\begin{bmatrix}\B{q}_{i} \\ 1\end{bmatrix}-\frac{1}{z_{i}}\B{I}_{3\times 4}\B{R}^{-1}\begin{bmatrix}\B{p}_{i} \\ 1\end{bmatrix}\right\Vert$$

Letting $\begin{bmatrix}\B{q}_{i}'\\ 1\end{bmatrix} = \B{R}_{\gamma}\begin{bmatrix}\B{q}_{i}\\ 1\end{bmatrix}$, we will thus minimize the reprojection error using the $\B{q}_{i}'$ while assuming that $\gamma = 0$. This assumption will simplify the computation of $\text{Error}(\B{R})$ and its derivatives. Assume that the camera has 2D pose $(x,y,\theta)$ relative to the world origin. Then, letting $\Delta x_{i} = x-p_{i,x}$ and $\Delta y_{i}=y-p_{i,y}$, we find:

$$\B{P} = \begin{bmatrix}\cos(\theta) & -\sin(\theta) & 0 & x \\ \sin(\theta) & \cos(\theta) & 0 & y \\ 0 & 0 & 1 & 0 \\ 0 & 0 & 0 & 1\end{bmatrix}\Rightarrow\begin{bmatrix}\B{R}_{\beta}^{-1} & \B{0} \\ \B{0}^{\intercal} & 1\end{bmatrix}\B{P}^{-1}\begin{bmatrix}\B{p}_{i}\\ 1\end{bmatrix}$$
$$=\begin{bmatrix}-(\Delta x_{i}\cos(\theta)+\Delta y_{i}\sin(\theta))\cos(\beta)-p_{i,z}\sin(\beta)\\ \Delta x_{i}\sin(\theta)-\Delta y_{i}\cos(\theta) \\-(\Delta x_{i}\cos(\theta)+\Delta y_{i}\sin(\theta))\sin(\beta)+p_{i,z}\cos(\beta)\\ 1\end{bmatrix}$$
$$\Rightarrow \text{Error}(\B{R})=\sum_{i}\left\Vert\begin{bmatrix}\B{q}_{i}' \\1\end{bmatrix}-\frac{1}{z_{i}}\B{I}_{3\times 4}\begin{bmatrix}\B{R}_{\beta}^{-1} & \B{0} \\ \B{0}^{\intercal} & 1\end{bmatrix}\B{P}^{-1}\begin{bmatrix}\B{p}_{i}\\ 1\end{bmatrix}\right\Vert^{2}$$
$$=\sum_{i}\left\Vert\begin{bmatrix}q_{i,x}' - \frac{(\Delta x_{i}\cos(\theta)+\Delta y_{i}\sin(\theta))\cos(\beta)+p_{i,z}\sin(\beta)}{(\Delta x_{i}\cos(\theta)+\Delta y_{i}\sin(\theta))\sin(\beta)-p_{i,z}\cos(\beta)}\\[6pt] q_{i,y}' - \frac{\Delta y_{i}\cos(\theta)-\Delta x_{i}\sin(\theta)}{(\Delta x_{i}\cos(\theta)+\Delta y_{i}\sin(\theta))\sin(\beta)-p_{i,z}\cos(\beta)}\end{bmatrix}\right\Vert^{2}$$

Because minimizing $\text{Error}(\B{R})$ is a nonlinear least-squares problem, we employ iterative optimization algorithms such as Levenberg-Marquardt (used by our implementation) or Gauss-Newton. These algorithms require the Jacobian matrix $\B{J}$, which can be found analytically using the above expression for $\text{Error}(\B{R})$. 

For an iterative optimization algorithm to consistently reach the correct local minimum, it must be provided with an appropriate initial guess. We describe a two-phase initialization strategy that takes all points into account. Define:

$$\B{R}_{\beta}\begin{bmatrix}\B{q}_{i}'\\ 1\end{bmatrix} = r_{i}\begin{bmatrix}\cos(\theta_{i})\cos(\phi_{i})\\ \sin(\theta_{i})\cos(\phi_{i})\\ \sin(\phi_{i})\end{bmatrix}$$

Here, $\theta_{i}$ and $\phi_{i}$ are the azimuth and elevation angles of $\B{q}_{i}'$ with respect to the camera's origin. Let $\Delta\theta_{i}$ and $\Delta\phi_{i}$ be the differences in azimuth and elevation between $\B{p}_{i}$ and $\B{q}_{i}'$ with respect to the camera's origin. This is illustrated in the following diagram:
\begin{figure}[b]
    \centering
    \captionsetup{justification=centering}
    \caption{Definitions of $\Delta\theta_{i}$ and $\Delta\phi_{i}$}
    \vspace{-1cm}
    \tdplotsetmaincoords{60}{45}
    \begin{tikzpicture}[scale=2,tdplot_main_coords, dot/.style = {circle, fill, minimum size=#1, inner sep=0pt, outer sep=0pt}]
    \begin{scope}[thick]
        \draw[dashed, blue] (3.86, 1.04, 0) -- (0, 0, 0) -- (2.7, 0.73, 1.5);
        \draw[dotted, blue] (2.7, 0.73, 0) -- (2.7, 0.73, 1.5) node[dot=4, label=-45:{$\B{q}_{i}'$}]{};
        \draw[dotted, green!75!black] (3.7, 1.53, 0) --(3.7, 1.53, 3) node[dot=4,label=180:{$\B{p}_{i}$}]{};
        \tdplotdrawarc[dotted, green!75!black]{(0,0,3)}{4}{15}{22.5}{}{};
        \draw[dashed,green!75!black](3.86, 1.04, 3) -- (0,0,0)--(3.7,1.53,0);
        \draw[dashed, black] (0, 0, 0) -- (4, 0, 0);
        \draw (3.5, 0) arc (0:15:3.5) node[pos=0.3,label={[label distance=2mm]right:{$\theta_{i}$}}]{};
        \draw (3.62, 0.97, 0) arc (15:22.5:3.75) node[pos=0.2,label={[label distance=1mm]right:{$\Delta\theta_{i}$}}]{};
        \tdplotsetrotatedcoords{0}{-30}{0};
        \draw[thick, tdplot_rotated_coords](3, 1.5, 1) -- (3, -1.5, 1) node[midway, rotate=25, above]{Image Plane}-- (3, -1.5, -1) -- (3, 1.5, -1) -- (3, 1.5, 1);
        \draw[densely dotted, tdplot_rotated_coords](3, 1.5, 1) -- (0, 0,0) node[dot=4, fill=black, label=135:{$(x,y)$}]{} -- (3, -1.5, 1);
        \draw[densely dotted, tdplot_rotated_coords](3, 1.5, -1) -- (0, 0, 0) -- (3, -1.5, -1);
        \tdplotsetrotatedcoords{-75}{90}{90};
        \tdplotdrawarc[tdplot_rotated_coords]{(0, 0, 0)}{2}{0}{28.2}{shift={(3mm,-1mm)}}{$\phi_{i}$};
        \tdplotdrawarc[tdplot_rotated_coords]{(0,0,0)}{2.75}{28.2}{36.8}{shift={(4mm,1.5mm)}}{$\Delta\phi_{i}$};
    \end{scope}
    \end{tikzpicture}
\end{figure}

Roughly speaking, a good initial pose will have small values for $\text{E}_{\phi} = \displaystyle\sum_{i}\Delta\phi_{i}^{2}$ and $\text{E}_{\theta} = \displaystyle\sum_{i}\Delta\theta_{i}^{2}$. Note that each $\Delta\phi_{i}$ does not depend on $\theta$. Thus, our overall strategy will be to find $\displaystyle\min_{x,y} \text{E}_{\phi}$ and then $\displaystyle\min_{\theta}\text{E}_{\theta}$. Although minimizing with respect to all variables at once is theoretically preferable, reasonable choices of objective function seem to make the problem analytically intractable. Thus, we would run into the same problem we are attempting to avoid - an iterative optimization algorithm becoming stuck in the wrong local minimum.

Note that $\displaystyle\cot^{2}(\phi_{i}+\Delta\phi_{i}) = \frac{\Delta x_{i}^{2} + \Delta y_{i}^{2}}{p_{i,z}^{2}}$. Thus, linearizing $\cot^{2}$ around $\phi_{i}$, we find:

$$\text{E}_{\phi}\approx\sum_{i}\left(\frac{\Delta x_{i}^{2}+\Delta y_{i}^{2}}{p_{i,z}^{2}}\frac{\sin^{3}(\phi_{i})}{\cos(\phi_{i})} - \sin(\phi_{i})\cos(\phi_{i})\right)^{2}$$

Here, we discard terms where $p_{i,z}$ and $\phi_{i}$ have opposite signs because linearization will fail. Finding the stationary points of $\text{E}_{\phi}$ is equivalent to finding the roots of system of two cubic polynomials, $\displaystyle A(x,y)=\frac{\partial \text{E}_{\phi}}{\partial x}$ and $\displaystyle B(x,y)=\frac{\partial \text{E}_{\phi}}{\partial y}$. We note that not all coefficients are linearly independent, so the form of the system is as follows:

\begin{align*}A&=c_{1}x^{3}+c_{1}xy^{2}+3c_{2}x^{2}+2c_{3}xy+c_{2}y^{2}+a_{1}x+c_{4}y + a_{2}\\B&=c_{1}y^{3}+c_{1}xy^{2} + 3c_{3}y^{2}+2c_{2}xy + c_{3}x^{2} + b_{1}y + c_{4}x + b_{2}\end{align*}

Many PnP methods involve solving a polynomial system; accordingly, advanced techniques such as Gröbner bases are often brought to bear \cite{OPnP}. In this case, basic theory will suffice: the system has a solution at $x$ if and only if the resultant $\text{Res}(A,B,y)$, computed as the determinant of the polynomials' Sylvester matrix, is zero at $x$. In general, because each polynomial has total degree $3$, the resultant's degree is at most $9$. Because of the linear dependence of the system's coefficients, however, we find that all terms above degree $5$ vanish. Although this is still out of the realm of algebraic solution, we can find all roots efficiently by finding the eigenvalues of the resultant's companion matrix \cite{NR}. Both polynomials in the system have relatively small degrees, so it is feasible to store the full expressions for each of the resultant's coefficients (amounting to a few thousand multiplications). 

Once we find the possible values of $x$, the usual method is to back-substitute into the original system to find the corresponding values of $y$. However, this may lead to numerical instability, so we elect to compute and fully solve the second resultant $\text{Res}(A,B,x)$ instead. Because each resultant has at most $5$ real roots, there are at most $25$ solution pairs to check (and usually fewer than $5$), which is not a significant computational penalty over back-substitution.

After we compute the possible positions, the next step is to find the optimal heading for each position. As an approximation to minimizing $\text{E}_{\theta}$, we maximize $\displaystyle\sum_{i}\cos(\Delta\theta_{i})$. This allows the following simplification by harmonic addition:

$$\sum_{i}\cos(\Delta \theta_{i}) = \sum_{i}\cos(\theta+\theta_{i}-\text{atan}_{2}(\Delta y_{i},\Delta x_{i}))$$
$$=A\cos(\theta + \delta),\ \tan(\delta) = \sum_{i}\frac{\Delta x_{i}\sin(\theta_{i}) - \Delta y_{i}\cos(\theta_{i})}{\Delta x_{i}\cos(\theta_{i}) + \Delta y_{i}\sin(\theta_{i})}$$

Thus, the possible optimal values of $\theta$ are $-\delta$ and $\pi-\delta$, depending on the values of $x$ and $y$ found in the previous step. Over all choices of initial pose (at most 50 and usually fewer), we take the pose with the lowest reprojection error and use it to initialize the iterative phase of the algorithm. We see that both phases have $O(n)$ time complexity.

\section{Alternative Initialization Strategies}
Other initialization strategies were also explored; for instance, a previous version of our algorithm chose only three far-apart correspondences to calculate a simple closed-form solution. However, this was prone to noise and ambiguity, resulting in higher overall failure rates. Applying the weak-perspective assumption as in \cite{LHM} can create a completely linear problem, but this assumption fails if the points are sufficiently widely distributed in camera space. Finally, applying one of the faster 3D PnP algorithms and simply projecting and polishing the solution in 2D was discounted for performance reasons, as well as the increased potential for ambiguous pose estimates with planar configurations of points. In summary, our proposed initialization strategy was chosen for its ability to consider all correspondences while remaining performant. 

If other, possibly approximate information concerning the camera's pose is known, the initialization strategy can be simplified greatly. For instance, if an approximate heading $\theta$ is provided by an IMU, we can find an initial position by minimizing the following error over $x,y$:

$$\text{E}_{xy} = \sum_{i}(x_{i}\sin(\theta+\theta_{i}) - y_{i}\cos(\theta+\theta_{i}))^{2}$$

This is a linear least-squares problem, which is easily solved in closed form. Unlike $\text{E}_{\theta}$, $\text{E}_{xy}$ does not measure the angular deviations between the object and image points; rather, it measures the distances between the camera rays and the object points when both are projected onto the $xy$-plane. As noted in \cite{EPnP}, the most important component of the initial pose is the heading, so this small deviation from the ``ideal" $\text{E}_{\theta}$ is unlikely to have an impact.
\begin{figure*}[b]
    \centering
    \captionsetup{justification=centering}
    \captionsetup[subfigure]{labelformat = empty}
    \begin{subfigure}{0.5 \linewidth}
        \includegraphics[width=\linewidth]{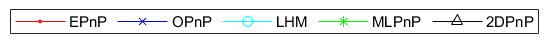}
    \end{subfigure}

    \vspace{0.75cm}
    \begin{subfigure}[b]{0.38 \linewidth}
        \includegraphics[width=\linewidth]{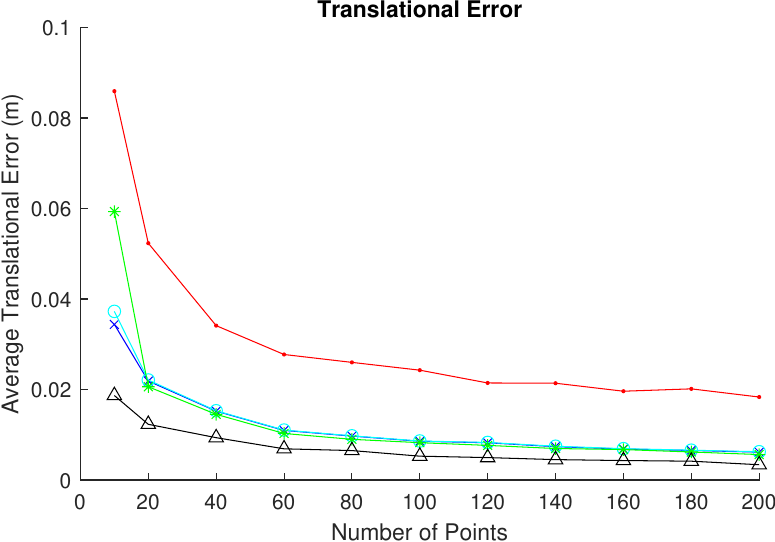}
        \caption{(a) Average translational error vs. number of points}
    \end{subfigure}
    \hspace{0.5cm}
    \begin{subfigure}[b]{0.38 \linewidth}
        \includegraphics[width=\linewidth]{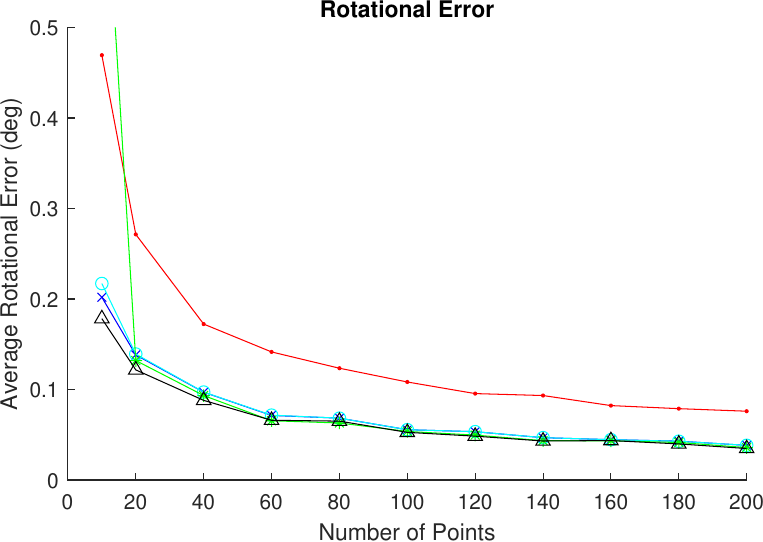}
        \caption{(b) Average rotational error vs. number of points}
    \end{subfigure}

    \vspace{0.5cm}
    \begin{subfigure}{0.8\linewidth}
        \includegraphics[width =\linewidth]{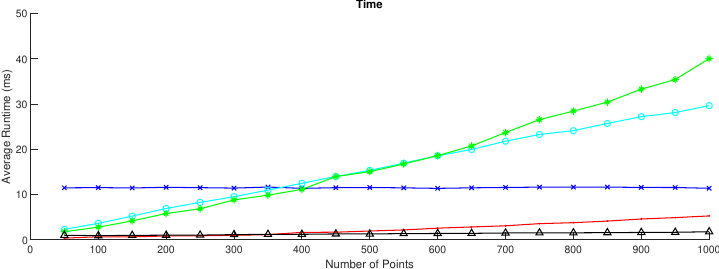}
        \caption{(c) Average run time vs. number of points}
    \end{subfigure}

    \vspace{0.5cm}
    \begin{subfigure}[b]{0.38 \linewidth}
        \includegraphics[width=\linewidth]{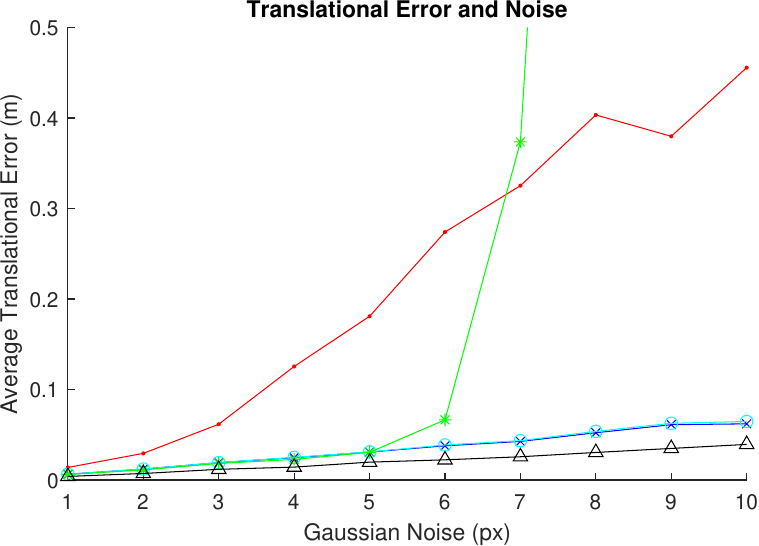}
        \caption{(d) Average translational error vs. amount of noise}
    \end{subfigure}
    \hspace{0.5cm}
    \begin{subfigure}[b]{0.38 \linewidth}
        \includegraphics[width=\linewidth]{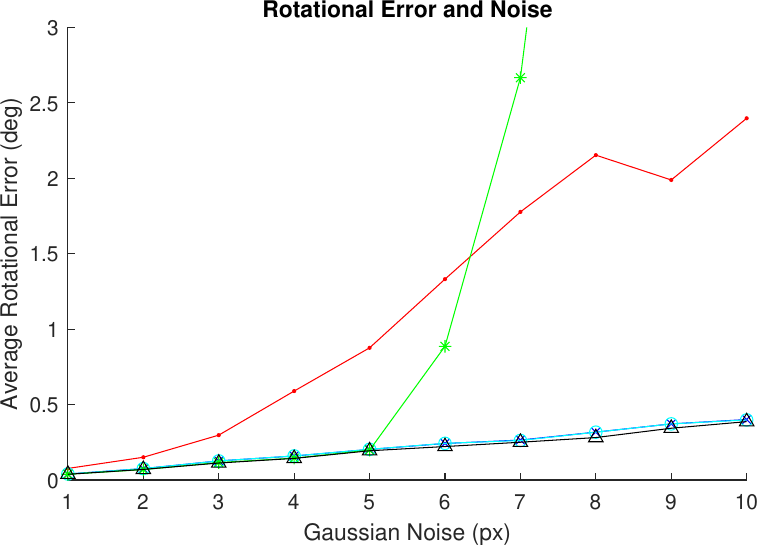}
        \caption{(e) Average rotational error vs. amount of noise}
    \end{subfigure}
\end{figure*}
\section{Testing}
We implemented the 2DPnP algorithm in MATLAB and tested it alongside existing PnP algorithms calculating 3D pose. For each test, points were randomly generated in the cuboid $[-2,2]\times [-2,2]\times[4,8]$ in camera space, projected onto the image plane assuming a focal length of 800 pixels, and each image point was perturbed with Gaussian noise. The camera's ground truth pose was $(0,0,0)$, and its rotational offset was a random unit quaternion within the range of pitch angles $\beta\in [10^{\circ},170^{\circ}]$; this avoids inherent singularities in the conversion of 3D rotation to 2D heading. For each algorithm returning a 3D pose, the pose was projected onto the $xy$-plane. Each data point represents the average of 250 independent tests. This follows the methodology of \cite{MLPnP}, and the 3D PnP implementations we tested were provided by their MATLAB toolbox.

We tested the average translational and rotational errors of each algorithm for numbers of points from 10 to 200. Gaussian noise with a standard deviation of $2$ pixels was applied to each image point. The same conditions were used to time each algorithm for numbers of points from 50 to 1000.

We also tested the impact of noise on each algorithm. For these tests, we used 50 points, and each image point was perturbed by Gaussian noise with a standard deviation from 1 to 10 pixels. 

The tested algorithms were \B{2DPnP} (ours), \B{LHM} \cite{LHM}, \B{EPnP} with Gauss-Newton refinement \cite{EPnP}, \B{OPnP} \cite{OPnP}, and \B{MLPnP} without covariance information \cite{MLPnP}. This is by no means a comprehensive list; because none of the 3D algorithms can accommodate for constraints to 2D motion, they cannot not be expected to perform optimally, and our intent is only to show how our method compares to a variety of approaches to the 3D problem. All tests were performed on a Intel i5 processor at 2 GHz. The raw data, plots, and MATLAB files are available at \href{https://github.com/25wangj/2DPnPToolbox}{\tt{https://github.com/25wangj/2DPnPToolbox}}.\newpage

\section{Conclusion}    
We presented a PnP algorithm for a camera constrained to two-dimensional motion. The algorithm, 2DPnP, calculates two resultants to solve a polynomial system for an approximate initial pose and applies Levenberg-Marquardt optimization refine the pose through minimization of the reprojection error. We tested 2DPnP against currently used PnP algorithms which find three-dimensional pose. Our algorithm demonstrated significantly better performance due to the reduced dimensionality of the search space, despite being implemented purely in MATLAB with little optimization. It also showed increased translational and comparable rotational accuracy, and increased robustness to noise. We believe that this algorithm can be fruitfully applied to the vision-based localization of wheeled mobile robots (prominent in educational and warehouse robotics), using either a camera onboard a robot or an external camera recognizing fiducial markers on each robot. It is especially promising in cases of limited resolution or computational power.

Further work could include the extension of the algorithm to multiple rigidly constrained cameras, enabling stereo or externally synchronized cameras to be utilized to their full potential. Additionally, more sophisticated initialization strategies may succeed in taking into account all three components of the pose at once, as opposed to the stepwise optimization paradigm that we currently use. Finally, approaches to PnP entirely distinct from the minimization of reprojection error have shown success in the general 3D problem; it is conceivable that these methods, such as analytically minimizing an algebraic error by solving a polynomial system \cite{OPnP} or finding a pose according to maximum-likelihood estimation \cite{MLPnP} could be applied to the 2D case. The latter is especially attractive because it can rigorously propagate image point uncertainty to camera pose uncertainty and may thus be used for more robust sensor fusion.
\bibliographystyle{ieeetr}
\bibliography{bibliography.bib}

\begin{thebibliography}{10}

\bibitem{P3P}
L.~Kneip, D.~Scaramuzza, and R.~Siegwart, ``A novel parametrization of the perspective-three-point problem for a direct computation of absolute camera position and orientation,'' in {\em CVPR 2011}, pp.~2969--2976, 2011.

\bibitem{P4P}
M.~Bujnak, Z.~Kukelova, and T.~Pajdla, ``A general solution to the p4p problem for camera with unknown focal length,'' in {\em 2008 IEEE Conference on Computer Vision and Pattern Recognition}, pp.~1--8, 2008.

\bibitem{LHM}
C.-P. Lu, G.~Hager, and E.~Mjolsness, ``Fast and globally convergent pose estimation from video images,'' {\em IEEE Transactions on Pattern Analysis and Machine Intelligence}, vol.~22, no.~6, pp.~610--622, 2000.

\bibitem{EPnP}
V.~Lepetit, F.~Moreno-Noguer, and P.~Fua, ``Epnp: An accurate o(n) solution to the pnp problem,'' {\em International Journal of Computer Vision}, vol.~81, 02 2009.

\bibitem{OPnP}
Y.~Zheng, Y.~Kuang, S.~Sugimoto, K.~Åström, and M.~Okutomi, ``Revisiting the pnp problem: A fast, general and optimal solution,'' in {\em 2013 IEEE International Conference on Computer Vision}, pp.~2344--2351, 2013.

\bibitem{IPPE}
T.~Collins and A.~Bartoli, ``Infinitesimal plane-based pose estimation,'' {\em International Journal of Computer Vision}, vol.~109, pp.~252--286, 2014.

\bibitem{Planar1}
P.-C. Wu, Y.-H. Tsai, and S.-Y. Chien, ``Stable pose tracking from a planar target with an analytical motion model in real-time applications,'' 11 2014.

\bibitem{Planar2}
G.~Schweighofer and A.~Pinz, ``Robust pose estimation from a planar target,'' {\em IEEE Transactions on Pattern Analysis and Machine Intelligence}, vol.~28, pp.~2024--2030, 2006.

\bibitem{EKF}
M.~A. Mehralian and M.~Soryani, ``Ekfpnp: extended kalman filter for camera pose estimation in a sequence of images,'' {\em IET Image Processing}, vol.~14, no.~15, pp.~3774--3780, 2020.

\bibitem{NR}
W.~H. Press, S.~A. Teukolsky, W.~T. Vetterling, and B.~P. Flannery, {\em Numerical Recipes 3rd Edition: The Art of Scientific Computing}.
\newblock USA: Cambridge University Press, 3~ed., 2007.

\bibitem{MLPnP}
S.~Urban, J.~Leitloff, and S.~Hinz, ``Mlpnp – a real-time maximum likelihood solution to the perspective-n-point problem,'' {\em ISPRS Annals of the Photogrammetry, Remote Sensing and Spatial Information Sciences}, vol.~III–3, p.~131–138, June 2016.

\end{thebibliography}
\end{document}